\def\paperID{1442}
\def\confName{CVPR}
\def\confYear{2024}

\def\paperTitle{Active Object Detection with Knowledge Aggregation and Distillation \\from Large Models}

\def\authorBlock{
    Dejie Yang \qquad
    Yang Liu\thanks{Corresponding Author} \\
    Wangxuan Institute of Computer Technology, Peking University\\
    {\tt\small ydj@stu.pku.edu.cn yangliu@pku.edu.cn}
}

\newif\ifreview 
\newif\ifarxiv 
\newif\ifcamera \newcommand{\cameraready}{\cameratrue}
\newif\ifrebuttal

\cameraready %\review  % \review OR \arxiv OR \cameraready

\pdfoutput=1
\documentclass[10pt,twocolumn,letterpaper]{article}
\ifreview \usepackage[review]{cvpr} \fi
\ifarxiv \usepackage[pagenumbers]{cvpr} \fi
\ifrebuttal \usepackage[rebuttal]{cvpr} \fi
\ifcamera \usepackage{cvpr} \fi

\usepackage{graphicx}	
\usepackage{amsmath}	
\usepackage{amssymb}	
\usepackage{booktabs}
\usepackage{times}
\usepackage{microtype}
\usepackage{epsfig}
\usepackage[table,xcdraw,dvipsnames]{xcolor}
\usepackage{caption}
\usepackage{float}
\usepackage{placeins}
\usepackage{color, colortbl}
\usepackage{stfloats}
\usepackage{enumitem}
\usepackage{tabularx}
\usepackage{xstring}
\usepackage{multirow}
\usepackage{xspace}
\usepackage{url}
\usepackage{subcaption}
\usepackage{xcolor}
\usepackage{graphbox}
\usepackage[hang,flushmargin]{footmisc}

\ifcamera \usepackage[accsupp]{axessibility} \fi

% ...

\ifarxiv  \fi

\newcommand{\R}[1]{{%
    \textbf{%
        \ifstrequal{#1}{1}{\textcolor{red}{R#1}}{%
        \ifstrequal{#1}{2}{\textcolor{blue}{R#1}}{%
        \ifstrequal{#1}{3}{\textcolor{magenta}{R#1}}{%
        \ifstrequal{#1}{4}{\textcolor{teal}{R#1}}{%
                           \textcolor{cyan}{R#1}%
        }}}}%
    }%
}}
\usepackage{xr-hyper}

\makeatletter
\newcommand*{\addFileDependency}[1]{
  \typeout{(#1)}
  \@addtofilelist{#1}
  \IfFileExists{#1}{}{\typeout{No file #1.}}
}

\makeatother

\definecolor{cvprblue}{rgb}{0.21,0.49,0.74}
\usepackage[pagebackref,breaklinks,colorlinks,citecolor=cvprblue]{hyperref}
\usepackage[accsupp]{axessibility}
\usepackage[capitalize]{cleveref}
\crefname{section}{Sec.}{Secs.}
\crefname{table}{Table}{Tables}
\crefname{figure}{Fig.}{Figs.}

\begin{document}
%% TITLE
\title{\paperTitle}
\author{\authorBlock}
\maketitle

\begin{abstract}
Accurately detecting active objects undergoing state changes is essential for comprehending human interactions and facilitating decision-making. The existing methods for active object detection (AOD) primarily rely on visual appearance of the objects within input, such as changes in size, shape and relationship with hands. However, these visual changes can be subtle, posing challenges, particularly in scenarios with multiple distracting no-change instances of the same category. We observe that the state changes are often the result of an interaction being performed upon the object, thus propose to use informed priors about object related plausible interactions (including semantics and visual appearance) to provide more reliable cues for AOD. Specifically, we propose a knowledge aggregation procedure to integrate the aforementioned informed priors into oracle queries within the teacher decoder, offering more object affordance commonsense to locate the active object. To streamline the inference process and reduce extra knowledge inputs, we propose a knowledge distillation approach that encourages the student decoder to mimic the detection capabilities of the teacher decoder using the oracle query by replicating its predictions and attention. Our proposed framework achieves state-of-the-art performance on four datasets, namely Ego4D, Epic-Kitchens, MECCANO, and 100DOH, which demonstrates the effectiveness of our approach in improving AOD. The code and models are available
at \url{https://github.com/idejie/KAD.git}.
\end{abstract}
\section{Introduction}
\label{sec:intro}

% To insert a figure: \input{figs/template}
% Or table: \input{tables/template}

\begin{figure}
  \centering
  \begin{subfigure}{0.48\linewidth}
    \includegraphics[align=c,width=1\linewidth]{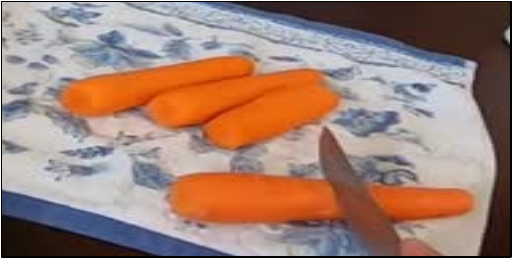}
    \caption{subtle difference and multiple distractors }
    \label{fig:teaser-a}
  \end{subfigure}
  % \\
 \hfill
  \begin{subfigure}{0.48\linewidth}
    \includegraphics[align=c,width=1\linewidth]{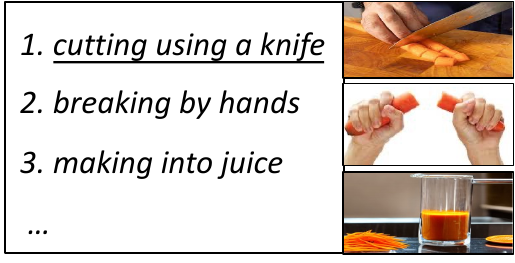}
    
    \caption{diverse interaction and large intra-class variance}
    \label{fig:teaser-b}
  \end{subfigure}

   \label{fig:teaser}
  \caption{\textbf{An example of state-change carrot.}  Active objects detection (state change carrots) is difficult, as there are (1) visual changes can be subtle between the carrot undergoing state-change or not, and multiple distractors, (2) intra-class visual appearance variance for the carrot under state changes is large. To achieve accurate detection, we propose to construct triple priors to provide hints for the model, including semantic interaction priors, fine-grained visual priors, and spatial priors of active objects. }
\end{figure}

Active object detection (AOD) focuses on localizing key objects that are undergoing state changes as a result of a sequence of human actions, interactions, or manipulations, which has broad potential applications \cite{grauman2022ego4d,fu2022sequential,han2019active}. 

Existing methods explore conventional object detectors, i.e., Faster RCNN~\cite{ren2015faster}, DETR~\cite{carion2020end}, CenterNet~\cite{duan2019centernet}, or specialized hand-object interaction models~\cite{shan2020understanding,fu2021sequential} for AOD, which primarily rely on visual appearance of the objects \textit{within input}, such as changes in size, shape and its relationship with hands. However, only using visual cues is often inadequate for AOD due to the following reasons:
(1) \textit{The visual changes can be subtle between the instance undergoing state-change or not}, posing challenges, particularly in scenarios with multiple distracting no-change instances of the same category in Figure\ref{fig:teaser-a}, or the target active object suffers from hands occlusion. (2) \textit{The intra-class visual appearance variance for the same object under state changes is large}. As shown in Figure\ref{fig:teaser-b}, a state-change of the carrot can be caused by many interactions, such as `cutting using a knife', `breaking by hands' or `making into juice'. 

In practice, we notice that alterations in the object's state frequently stem from interactions carried out on the object. This underscores the significance of common-sense understanding of object affordance in the context of AOD.  If we can establish well-informed priors that model this common-sense knowledge concerning plausible object interactions beforehand (out of the scope of the current input), we can uncover the aforementioned `intra-class variance'. In principle, this revelation can be leveraged to assist AOD, analogous to leveraging hints during exams to facilitate question-solving.

Motivated by this, in this paper, we propose a new framework, namely `knowledge aggregation and distillation'(KAD), aiming at aggregating the object affordance common sense knowledge and incorporating it into the AOD process as shown in Figure\ref{fig:framework}. Firstly, we propose to model such  common sense in three ways :(1) we employ a language model to provide \textit{semantic interaction priors}  of ``how the active object can be interacted," capturing multiple possibilities that may induce changes in the object's state; (2) we use an image generation model to equip the above semantic by providing vivid images depicting state changes during corresponding interactions, offering \textit{fine-grained visual priors}; (3)we utilize the ground truth position of the active object as  \textit{spatial priors}, indicating regions that require heightened attention to enhance the model's spatial sensitivity, particularly in the presence of multiple distractors. Secondly, we introduce a Knowledge Aggregator designed to reconcile conflicts and harness complementary cues from the three aforementioned common-sense modalities, serving as an oracle query for the decoder. Thirdly, in practical scenarios, the category of active objects remains unknown during inference, rendering the acquisition of the aforementioned oracle query challenging. As a solution, we propose a distillation strategy that uses a plain detector as student (using learnable queries) to mimic the attention and intermediate outputs in a teacher detector using an oracle query. Throughout the training process, we facilitate the transfer of knowledge from the `cheated' teacher to the student through distillation, thereby endowing the student with a level of proficiency in active object detection ability akin to that of the teacher. During inference, we leverage the student detector exclusively, effectively circumventing the need for unnecessary inputs.

Our contributions can be summarised as follows:
\begin{itemize}
    \item We introduce a Knowledge Aggregator that incorporates three-fold commonsense pertaining to active objects, encompassing plausible semantic interactions, fine-grained visual and spatial priors, serving as a `cheated' teacher to facilitate more accurate AOD localization.
    \item To avoid the extra commonsense input at inference, we propose a Teacher-student Knowledge Distillation strategy, enabling the training of a simple student detector that possesses robust AOD capabilities by mimicing the attention and intermediate outputs from the teacher.
    \item Comprehensive experiments conducted on  extensive egocentric datasets, Ego4D, Epic-Kitchens, MECCANO and 100DOH, demonstrate the efficacy of our proposed approach.
\end{itemize}
\section{Related Work}
\label{sec:related}

\subsection{Active Object Detection}\label{AOD}
Active object detection (AOD) involves identifying manipulated objects entwined with human actions\cite{fu2021sequential,thakur2023anticipating,wang2022internvideo,dunnhofer2023visual}.
InternVideo\cite{wang2022internvideo} proposes a series of models for Ego4D tasks. For state-change object detection, InternVideo\cite{wang2022internvideo} explores the transfer learning
from general object detection to egocentric state-change object detection, which takes the Swin-Transformer\cite{liu2021Swin} as the backbone and employs DINO\cite{zhang2022dino} as the detection head.
HOTR\cite{kim2021hotr} proposes an end-to-end human or hand object interaction detection transformer, which detects hand(or human) and object independently with a detector and identifies the hand-object interaction to match hands and objects detected in the detector.
Seq-Voting\cite{fu2021sequential} takes the hand detection results as cues and proposes a voting function with a box field to leverage each pixel of the input image as evidence to predict the bounding box of the active object. While the investigation of hand-object interactions\cite{fu2021sequential,shan2020understanding,kim2021hotr} stands out as a crucial facet of object manipulation, it is crucial to recognize that objects undergoing state changes might not always be directly engaged with hands. 
Only using visual cues is often inadequate because of the subtle difference between state-change or not and the large intra-class visual appearance of the same active object.
In this paper, we aim to provide three-fold priors: semantic interaction priors, fine-grained visual priors, and spatial-sensitive priors to enhance active object detection.

\subsection{Object-State Change}
Human actions often induce changes of the state of an object. Previous works have studied detecting object states in images \cite{misra2017red,nagarajan2018attributes, gouidis2023leveraging, Saini_2023_ICCV} or learning actions and their modifiers in videos\cite{souvcek2022multi,souvcek2022look}. 
Gouidis et al.\cite{gouidis2023leveraging} propose an approach for the task of zero-shot state classification, which combines the knowledge graph on object states and visual information that relates the appearance of certain objects to their states.
ChopNLearn\cite{Saini_2023_ICCV} aims to learn compositional generalization, the ability to recognize and generate unseen compositions of objects in different states. 
\cite{wu2023localizing} aims to grounding the active objects, and proposes a  prompting pipeline to extract 
knowledge for objects undergoing state change.
Unlike these methods for object state grounding\cite{wu2023localizing}, recognition or generation\cite{misra2017red,nagarajan2018attributes, gouidis2023leveraging, Saini_2023_ICCV}, the AOD task aims to detect/locate objects undergoing state changes.
Locating active objects not only requires semantic and visual priors but also relies on spatial prior to provide clear hints to distinguish between active objects and other distractors. In this paper, we aim to construct an oracle query for the detector and the query contains semantic, visual, and spatial priors.

\subsection{Knowledge Distillation in Object Detection}
Traditional object detection entails the task of identifying and localizing all objects within an image or video frame, encompassing the simultaneous duties of classification and spatial localization. 
Recently, the successful application of knowledge distillation in traditional object detection has garnered attention\cite{li2023object,changdetrdistill,wang2019distilling,zhang2020improve,yang2022prediction}. In the pursuit of compact and efficient object detection networks, \cite{chen2017learning} have seamlessly integrated knowledge distillation, achieving heightened efficiency with minimal accuracy trade-offs. 
Additionally, \cite{kang2021instance} proposes an ingenious approach that incorporates instance annotations into an attention mechanism, effectively pinpointing significant regions.
FGD\cite{Yang_2022_CVPR} proposes a focal and global knowledge distillation to separate the objects and background and rebuild the relation between different pixels from teachers to students.
While knowledge distillation is effective in traditional object detection, its use in active object detection is novel. 
Attention or output-based distillation methods \cite{Yang_2022_CVPR,kang2021instance} in traditional object detection can provide certain spatial knowledge to distinguish between objects and backgrounds. However, AOD is more difficult, such as the subtle difference between state change or not.
It is inadequate to only provide spatial prior knowledge and distillation. Therefore, our method also integrates semantic interaction and fine-grained visual priors of active objects.

\section{Method}
\label{sec:method}
\begin{figure*}[h]
  \centering
   \includegraphics[width=\linewidth]{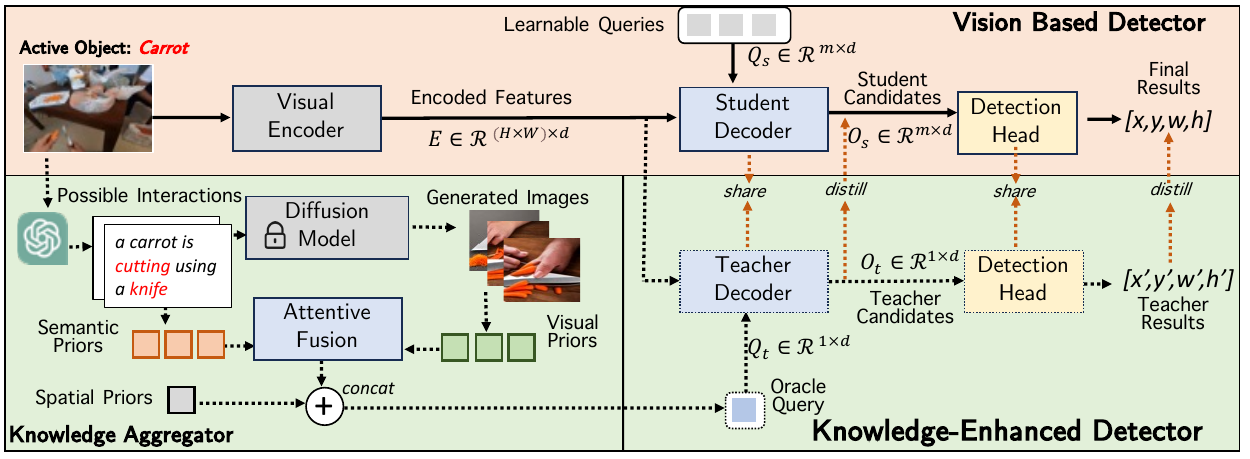}

\caption{\textbf{Proposed Architecture: Knowledge Aggregation and Distillation (KAD)}. Our KAD architecture comprises two distinct detectors: the Vision-Based Detector (highlighted in orange, detailed in Section \ref{subsec:overivew}) and the Knowledge-Enhanced Detector (emphasized in green, elaborated in Section \ref{subsec:KA}). Knowledge and concepts related to active object categories are systematically gathered and consolidated within the Knowledge Aggregator (shown in gray and positioned at the lower left, discussed in Section \ref{subsec:KD}). Best view in color. }
   \label{fig:framework}
\end{figure*}

\subsection{Overview of the pipeline}
\label{subsec:overivew}
As depicted in Figure \ref{fig:framework}, our newly devised Knowledge Aggregation and Distillation (KAD) framework encompasses two distinct detectors: the \textbf{Vision Based Detector} (emphasized in orange) and the \textbf{Knowledge-Enhanced Detector} (shaded in green). Moreover, we use \textbf{Knowledge Transfer} to enhance the detection capability of Vision Based Detector.

\textit{Vision Based Detector}: The detector is based on Transformer architecture, following \cite{carion2020end,zhang2022dino}. To extract the features of the frame/image, we adopt a visual backbone and a Transformer-based encoder to extract the feature map and encode the map. Then, a Transformer-based decoder is introduced to predict a set of candidates of active objects and model their relation. Finally, we use an active object detection head based on feed-forward networks to predict the normalized box of the active object and the confidence score of predictions.

\textit{Knowledge-Enhanced Detector}: To enrich the detection process with pertinent priors linked to active objects, we establish the Knowledge Aggregator, a component responsible for collecting the semantic-aware, visual-assisted and spatial-sensitive knowledge aligned with the active object's category (detailed in section \ref{subsec:KA}).
 And we use the encoded feature from vision based detector as input, and introduce a Transformer decoder  and an active object detection head  (both the components shared with vision based detector) to predict the normalized box of active object and the confidence score of prediction.

\textit{Knowledge Transfer:} 
Due to the unknown active object in the input image during testing, it is also difficult to obtain the relevant prior knowledge. 
To enhance the ability of active object detection for vision-based detector, we share the parameters of the decoders and detection heads of the above two detectors, and propose a knowledge distillation strategy to  transfer the knowledge from knowledge-enhanced detector to  vision-based detector by aligning the attention and immediate outputs.

\subsection{Vision Based Detector } 
\label{subsec:VBD}

In accordance with the conventional Transformer-based Detectors paradigm \cite{carion2020end,zhang2022dino}, we  use a visual backbone in tandem with a Transformer-based encoder to encode region features $\textbf{E} \in \mathbb{R}^{H \times W \times d}$ from the frame $\textbf{I}$.
In these expressions,  $H \times W$ shows the spatial resolutions of the  encoded features. The parameters $d$ represents the dimension of the encoded features.
Subsequently, a decoder furnished with a collection of learnable queries $\textbf{Q}_s \in \mathbb{R}^{m\times d}$ is introduced to forecast potential active object candidates, denoted as student candidates $\textbf{O}_s \in \mathbb{R}^{m\times d}$. In this context, $m$ corresponds to the number of learnable queries, which is equal to the number of student candidates.
Concluding this process, an active object detection head (depicted in yellow) outputs both the class-score $\hat{s} \in [0,1]$ (representing the confidence of active object detection) and the box location $\hat{b} \in \mathbb{R}^{m\times 4}$ for the student candidates.
To optimize the final predictions $(\hat{s}, \hat{b})$, we use the bipartite matching  following \cite{carion2020end} to find the lowest-cost pair between ground-truth $({s}, {b})$ (where $s=1$ due to only taking one active object annotation without inactive object annotations) and   $m$ predictions. Assume the $i$-th prediction $(\hat{s}_i , \hat{b}_i)$ is corresponding to lowest-cost $\mathcal{L}_{match}(i) =-\hat{s}_i+\lambda(\mathcal{L}_{giou}(b, \hat{b}_i)+||b- \hat{b}_i||_1)$, where $\mathcal{L}_{giou}$ is the Generalized IOU Loss and $||\cdot,\cdot||_1$ is the L1 distance, and $\lambda$ is the balanced parameter. Then we try to optimize the  $i$-th confidence score and box with the objective:
 \begin{equation}
     \mathcal{L}_v = BCE(s,\hat{s}_i)+\lambda(\mathcal{L}_{giou}(b, \hat{b}_i)+||b- \hat{b}_i||_1),
     \label{eq:l_V}
 \end{equation} where $BCE$ is the binary cross-entropy loss.

\subsection{Knowledge-Enhanced Detector} 
\label{subsec:KA}

As discussed in sections \ref{sec:intro} and \ref{subsec:overivew}, active object detection should consider  the subtle appearance changes (\textit{vision}), diverse possible interactions(\textit{semantic}), and distractions by other objects(\textit{spatial}). However, conventional vision-based detectors\cite{carion2020end,wang2022internvideo,Shan20,fu2021sequential} 
 overlook the significance of such common sense. Consequently, they lack the robustness required for accurate active object detection. Recognizing the potential of aggregating knowledge connected to active objects to furnish essential priors and cues, we introduce the \textbf{Knowledge Aggregator} in this section. Its primary role is to gather and fuse visual-assited, semantic-aware and spatial-sensitive knowledge related to active objects, thereby enriching the training phase with active-relevant priors.

Conversely, during the reasoning phase, predicting the category of active objects becomes inherently challenging, as they cannot always be effectively represented by the aforementioned related concepts. To address this, we present a \textbf{Knowledge Distillation} strategy. This approach compels vision-based detectors to align their intermediate outputs and attention mechanisms with those of knowledge-enhanced counterparts. 
By distillation, the student imitates the teacher's ability to detect AOD with priors and   avoids the extra commonsense input at inference.

\subsubsection{Knowledge Aggregator}
In order to overcome the challenge of AOD, the subtle appearance difference and large intra-class variance, we aim to aggregate the object affordance common sense knowledge and incorporating it into the AOD.
Specifically, we construct triple complementary priors: semantic interactions priors to capture multiple possibilities that may cause changes in object state, fine-grained visual priors to provide vivid images depicting state changes, and spatial priors to guided the model to distinguish where to pay more attention.

\textit{Semantic Interaction Priors.} To build an oracle query that includes active object related interactions, we use GPT to generate multiple descriptions of the scene where an object may be undergoing state change, for example: \textit{\textbf{carrot} is cutting using a knife}.
These describe concepts such as related objects and actions involved in the state changes of objects, and include relevant knowledge of active objects. We use language encoder to extract features of these descriptions {$[\mathbf{t}_1,\mathbf{t}_2,...,\mathbf{t}_p]$} to provide possible semantic  priors to build oracle query, where $p$ is the number of descriptions.

\textit{Fine-grained Visual Priors.} Semantic description of interaction is still abstract for vision-based detector. More directly, images  provide  more vivid visual information of related objects and interaction to better identify active objects. Therefore, we use the interaction description of the active object as prompt, and then use the Diffusion Model to generate the corresponding image. Similarly, we also extract the corresponding features from these images as a candidate set of visual concepts for the oracle query: {$[\mathbf{v}_1,\mathbf{v}_2,...,\mathbf{v}_q]$}, where $q$ is the number of images.

In order to select more important information from these text concepts and visual concepts, we propose an Attentive Fusion module to selectively aggregate these concepts. Specifically, taking text concepts as an example, we use the self-attention layer and max-pooling to select relatively important information:
\begin{equation}
    \textbf{T} = pool({selfattn([(\mathbf{t}_1,\mathbf{t}_2,...,\mathbf{t}_p)])}) \in \mathbb{R}^{1 \times d_t}.
    \label{eq:ag}
\end{equation} Similarly, we can also obtain selectively aggregated visual  priors: $\textbf{V} \in \mathbb{R}^{1 \times d_v}$. $d_t$ and $d_v$ are dimensions of the fused semantic prior and visual prior, respectively.

\textit{Spatial Priors.} AOD is also limited by the influence of other no change objects in the image, especially the objects with the same category of active object, which can easily distract attention in spatial. Active object is usually unique in an image.  It is also necessary to provide accurate spatial location  for AOD to  heighten attention to enhance the model's spatial sensitivity. We utilize the ground truth bounding box of the active object as {spatial priors}.

Finally, we merge text concepts, visual concepts and active object locations in the input image as the final oracle query, which has provided rich active object clues: $Q_t = [\textbf{T};\textbf{V};b]\in \mathbb{R}^{1 \times d}$, where $b \in \mathbb{R}^{1 \times 4}$ is the bounding box(spatial priors).

\subsubsection{Knowledge-Enhanced Detector}
\label{subsec:KD}
We strive to formulate an oracle query for active objects that encapsulates not only the associated knowledge but also the category and the normalized ground-truth bounding box of the active object. By harmonizing these three types of embeddings, we craft a comprehensive query denoted as $\textbf{Q}_t \in \mathbb{R}^{1 \times d}$.
Subsequently, the teacher decoder interfaces with the encoded features from the Vision-Based Detector, utilizing the amalgamated oracle query to provide critical semantic priors and precise positional data that serve as key indicators for detection. The resulting outputs from this teacher decoder, termed the Teacher Candidates $\textbf{O}_t \in \mathbb{R}^{1 \times d}$, are depicted in Figure \ref{fig:framework}.
Ultimately, this enhanced detector also employs detection heads to forecast the active object's presence, leveraging the insights gleaned from the oracle query $\textbf{Q}_t$
The optimization of the detector is similar to ${L}_v$(Eq.\ref{eq:l_V}):
 \begin{equation}
     \mathcal{L}_k = BCE(s,\hat{s}_t)+\lambda(\mathcal{L}_{giou}(b, \hat{b}_t)+||b- \hat{b}_t||_1),
     \label{eq:l_K}
 \end{equation} where $(\hat{s}_t \in \mathbb{R}^1,\hat{b}_t \in \mathbb{R}^{4})$ are the final active object result (confidence score and box corresponding to oracle query).

\subsection{Knowledge Distillation between Detectors}

The oracle query $\textbf{Q}_t$ integrates the three-fold informed priors for AOD: semantic interaction priors, fine-grained visual priors and spatial priors. Compared to Vision-Based Detector(student) using learnable queries, Knowledge-Enhanced Detector(teacher) using the oracle query can use the above priors to more accurately locate active objects.
Due to the unknown active object in the input image during testing, it is also difficult to obtain the relevant prior knowledge. We need to transfer the knowledge of the teacher to the student to avoid extra commonsense input and make student work well in alone at inference.

\textit{Parameter Sharing}. In order to achieve knowledge transfer from teacher to student, we share the parameters of decoders and detection heads between teacher and student.

\textit{Knowledge Distillation}. Furthermore, we adopt a distillation strategy from teacher to student, allowing the student to mimic the output and attention of the teacher. Specifically, Knowledge-Enhanced Detector takes encoded features $E$ as inputs and incorporates the oracle query to predict and learn the representation $O_t$ associated with an active object.
To augment the aptitude of the Vision-Based Detector for active object detection, we introduce a knowledge distillation mechanism between the predictions $\textbf{O}_t$ of the Knowledge-Enhanced Detector (teacher) and $\textbf{O}_{s}$ of the Vision-Based Detector (student).

Specifically, we align the cross attention $A_{s,i}$ and decoder embedding $O_{s,i}$  for $i$-th prediction of student network with those  ($A_t $ and $O_t$) of teacher via distillation:  
\begin{equation}
\begin{aligned}
L_{attn} = &\sum^{l} \mathrm{KL}(A_t^{l},A_{s_i}^l),\\
L_{emb} = \sum^{l}   & \left(1-  \frac{   {O_t^{l}}^T  O_{s_i}^{l}  }{  \Vert O_t^{l}\Vert _2\Vert O_{s_i}^{l} \Vert _2}\right).
\end{aligned}
\end{equation}
Here, $i$ corresponds to the index associated with the lowest-cost in the bipartite matching of the Vision-Based Detector. The parameter $l$ indicates the $l$-th layer in the decoders, which are shared between the two detectors. For a given $l$-th decoder layer, our approach involves aligning the cross attention pertaining to the $i$-th prediction in the student decoder, denoted as $A_{s_i}^l$, with its counterpart from the teacher, denoted as $A_t^l$, through a Kullback-Leibler divergence loss (KL). Additionally, we align the intermediate embeddings of the two networks using a cosine similarity loss. Aligning the embeddings forces the student to mimic the teacher's ability to express active objects. And attention can enable student to learn the teacher's ability where to pay attention to active objects in spatial.

The overall distillation loss is a combination of these two components, modulated by a hyper-parameter $\eta$, to achieve balance: $L_{distill} = L_{emb} + \eta L_{attn}$.

In essence, the strategy we employ begins by leveraging the oracle query to facilitate accurate representation learning in the teacher network, which helps address the challenges posed by dynamic distractors. Subsequently, we synchronize the intermediate outputs of the student network with those of the teacher network via distillation. This approach allows the student network to emulate the teacher network's ability to navigate dynamic distractors adeptly and to acquire the robust representation skills it possesses.

\subsection{Training and Inference}
\label{subsec:infer}
\quad \textit{Objective Functions}
The final objective function is as follows: 
\begin{equation}
\label{eq:loss}
\begin{aligned}
L =  L_v+ L_k + \alpha L_{distill}\  \text{,}
\end{aligned}
\end{equation}

\textit{Training and Inference}
During the training phase, the student detector's representation and attention mechanisms are structured to mimic those garnered from the teacher detector, thereby steering the student network via a distillation loss. This approach facilitates the transfer of valuable insights and knowledge from the teacher to the student, enhancing the student detector's aptitude.

Upon transitioning to the inference stage, the teacher detector is no longer in play. This strategic abandonment of the teacher model ensures that no supplementary computational overhead is incurred. 

\section{Experiment}
\label{sec:exp}

\begin{table*}[tb]
    \centering
         \setlength{\abovecaptionskip}{0cm} 
        \caption{Comparisons with other methods on Ego4D. We bold the best results and underline the second best ones. 
            }
\scalebox{0.9}{\begin{tabular}{c|c|c c c}
    
    \hline
    \multirow{2}{*}{Method}&\multirow{2}{*}{Backbone}& \multicolumn{3}{c}{Val-Set} \\
	         &  & AP&AP50&AP75 \\ \hline

        CenterNet~\cite{zhou2019objects} &DLA-34&  6.4&11.70&6.10 \\ \hline
       FasterRCNN~\cite{ren2015faster} &ResNet-101&  13.4&25.6&12.5 \\ \hline
               100DOH-model~\cite{Shan20}&ResNet-101 &  10.7 &20.6&10.1 \\ \hline
       DETR~\cite{carion2020end} &ResNet-50& \underline{15.5} &\underline{32.8}&\underline{13.0}\\ \hline
        KAD(ours) &ResNet-50&  \textbf{31.4}&\textbf{34.6}&\textbf{28.9}\\ \hline\hline
       \multirow{2}{*}{InternVideo\cite{wang2022internvideo}}&Uniformer-L&24.8&44.2&24.0\\ 
      &Swin-L&\underline{36.4}&\underline{56.5}&\underline{37.6}\\ \hline

       KAD(ours) &Swin-L& \textbf{40.5}&\textbf{60.6}&\textbf{41.9}\\ \hline

            \end{tabular}}
            
            \label{tab:ego4d}

\end{table*}

\begin{table*}[tb]
    \centering
         \setlength{\abovecaptionskip}{0cm} 
          \caption{Comparisons with other methods on Epic-Kitchens. We bold the best results and underline the second best ones. }
\scalebox{1}
{\begin{tabular}{c|c|c c c}
    
    \hline
    \multirow{2}{*}{Method}&\multirow{2}{*}{Backbone}& \multicolumn{3}{c}{Val-Set} \\
	         &  & AP&AP50&AP75 \\ \hline

       DETR~\cite{carion2020end} &ResNet-50&\underline{10.4} &\underline{15.7}&\underline{10.1}\\ \hline
       KAD(ours) &ResNet-50& \textbf{30.2}&\textbf{30.1}&\textbf{22.5} \\ \hline\hline
       \multirow{2}{*}{InternVideo\cite{wang2022internvideo}}&Uniformer-L&19.4&38.7& 17.0\\ 
      &Swin-L&\underline{28.3}&\underline{39.8}&\underline{27.2}\\ \hline

KAD(ours) &Swin-L& \textbf{35.2}&\textbf{44.1}&\textbf{32.5}\\ \hline

            \end{tabular}}
          
            \label{tab:epic}

\end{table*}

\begin{table*}[t]
    \centering
         \setlength{\abovecaptionskip}{0cm} 
        \caption{Comparisons with other methods on MECCANO. We bold the best results and underline the second best ones. }
\begin{tabular}{c|c|c c c}
    
    \hline
    \multirow{1}{*}{Method}&\multirow{1}{*}{Backbone}
	       & AP75&AP50&AP25 \\ \hline

       100DOH-model~\cite{Shan20}&ResNet-101 &  -&20.2&- \\ \hline
       
     Seq-Voting\cite{fu2021sequential} &ResNet-101&\underline{13.0}&\underline{26.3}&\underline{34.9}\\ \hline

  KAD(ours) &ResNet-101& \textbf {14.4}&\textbf {28.8}&\textbf {36.2} \\ \hline

            \end{tabular}
            
            \label{tab:MECCANO}

\end{table*}

\begin{table*}[t]
    \centering
    \setlength{\abovecaptionskip}{0cm} 
\caption{Comparisons with other methods on 100DOH. We bold the best results and underline the second best ones. }
\begin{tabular}{c|c|c c c}
    
    \hline
    \multirow{1}{*}{Method}&\multirow{1}{*}{Backbone}& AP75&AP50&AP25 \\ \hline

       100DOH-model~\cite{Shan20}&ResNet-101 &28.5&47.0&51.8 \\ \hline
       PPDM\cite{liao2020ppdm}& DLA-34&26.9&45.8&53.0\\ \hline
       HOTR\cite{kim2021hotr}&ResNet-50 &29.3&49.3&\underline{57.8}\\ \hline
     Seq-Voting\cite{fu2021sequential} &ResNet-101&\underline{29.9}&\underline{53.0}&57.2\\ \hline

  KAD(ours) &ResNet-101& \textbf {31.2}&\textbf {53.9}&\textbf {58.9} \\ \hline

            \end{tabular}

            \label{tab:100DOH}

\end{table*}

\subsection{Dataset}
\textbf{Ego4D}~\cite{ego4d} stands as one of the latest expansive egocentric video datasets. We focus on subsets of this dataset for our state-change object detection (SCOD) tasks. The original train and validation sets encompass 19,070 and 12,800 annotated frames, respectively, marking the point of no return, or the initiation of a state change. 

\noindent \textbf{Epic-Kitchens}~\cite{epic-kitchen} is a large-scale dataset in the domain of egocentric vision. We convert the segmentation annotations of action-related objects within the VISOR subset~\cite{VISOR2022} into bounding boxes, specifically tailored for the active object detection task. Notably, we employ a total of 67,217 and 9,668 annotated frames for our train and validation splits.

\noindent \textbf{MECCANO}~\cite{ragusa_MECCANO_2023} is an egocentric dataset for human-object interaction understanding in industrial-like settings. And it has been acquired in an industrial-like scenario in which subjects built a toy model of a motorbike. It contains 64,349 frames which are annotated with active object boxes.  Following prior splits\cite{fu2022sequential}, the training set, validation set, and test set contain 21686, 4270 and 12111 images, respectively.

\noindent \textbf{100DOH}~\cite{Shan20} is a large-scale benchmark for hand-object interaction. It has 99,899 frames (79,921 for training, 9,995 for validation and 9,983 for testing).  The focus of the dataset is hand contact, and it includes both first-person and third-person perspectives.
 
\subsection{Implementation Details}\label{detail}
 The embeddings of semantic and visual features are extracted through CLIP\cite{radford2021learning}. During training, we utilize AdamW optimization, setting the initial learning rate of the Transformer to $10^{-4}$, and the learning rate for the backbone to $10^{-5}$.
The hyperparameters $\alpha$, $\lambda$, and $\eta$ are configured to $0.2$,  $5.0$, and $1.0$, respectively, to govern the optimization process. Our models are trained over 50 epochs using a cosine annealing strategy with warm restarts. The training process is executed across  NVIDIA A100, and employs a batch size of 4.

\subsection{Comparisons Results}

\textbf{Comparisons on Ego4D and Epic-Kitchen}  
We present the COCO-style Average Precision (AP), AP50, and AP75 results achieved by our method on the Ego4D validation and test sets. The comparative outcomes for Ego4D are meticulously tabulated in Table~\ref{tab:ego4d}. (1) Our KAD method attains state-of-the-art performance across all metrics under a fair comparison protocol (with two backbones). This indicates the generalization of our model on different backbones. (2) From the Table~\ref{tab:ego4d}, it can be seen that the transformer based method can achieve higher performance. In particular, KAD outperforms the best existing method\cite{wang2022internvideo} with the same {transformer-based} backbone  by 4.1\%, 4.1\%, and 4.3\% on AP, AP50, and AP75, respectively.
This proves the effectiveness of priors for active object detection.  Notably, our method dose not introduce extra priors during testing(traditional detectors\cite{carion2020end,wang2022internvideo}), but instead forced traditional detectors to learn the AOD capabilities brought by priors through parameter sharing and knowledge distillation.

Table \ref{tab:epic} presents a comprehensive comparison between our KAD approach and other methods on the Epic-Kitchens dataset\cite{epic-kitchen}. To gauge the performance, we employed the DETR\cite{carion2020end} and InternVideo\cite{wang2022internvideo} methods on the Epic-Kitchens dataset, retraining and evaluating their outcomes using the corresponding configurations (by 6.9\%, 4.3\%, and 5.3\% on AP, AP50, and AP75 compared to best baseline method\cite{wang2022internvideo} ).
Importantly, our KAD method outperforms other approaches across all metrics. This improvement of performance underscores the effectiveness of our proposed KAD framework and its generalization ability on different datasets. 

\textbf{Comparisons on MECCANO and 100DOH} 
Due to the lack of object categories provided in 100DOH and fine-grained toy components in MECCANO\footnote{i.e., \textit{gray angled perforated bar}.}, it is difficult to generate semantic aware interaction descriptions and visual assigned images.
Therefore, we only use the spatial location of the active object as the oracle query. (1)From Tables \ref{tab:MECCANO} and \ref{tab:100DOH}, it can be seen that although our method only includes spatial cues, it still improves the AOD detection performance {without leveraging any external knowledge, which indicates the effectiveness of our network design}.  On the MECCANO dataset~\cite{ragusa_MECCANO_2023}, our method outperforms the previous best method\cite{fu2022sequential} in terms of AP75, AP50 and AP25 improved by 1.4\%,2.5\% and 1.3\% respectively. On 100DOH~\cite{Shan20},  our method improves the detection performance by 1.3\%, 0.9\%, and 1.7\%, respectively.It can be seen that spatial cues help the model focus on active objects and improve detection performance. (2) These two tables contain comparisons with more AOD specific methods {which considering the interactiveness}. And compared to them, our improvement indicates that only hand or visual information is not sufficient,  as well as the effectiveness on spatial priors.

\subsection{Ablation Study}
\label{ab}
In this part, we evaluate the effectiveness of different modules or variants of our KAD  on the Ego4D validation set.

\begin{table}
  \centering
  \setlength{\abovecaptionskip}{0cm} 
   \setlength{\abovecaptionskip}{0cm} 
        \setlength{\belowcaptionskip}{-0.2cm}
  \caption{Different  knowledge aggregations on Ego4D.  }
        \scalebox{0.9}
{
  \begin{tabular}{@{}c|c|ccc@{}}
    \toprule
    {No.}&{Knowledge}& AP&AP50&AP75\\
    \midrule

   1&  VBD(baseline)&{35.9}&{55.8}&{36.9}\\
   2&  VBD+visual&36.0 &56.6 &37.2\\
   3&  VBD+semantic&36.5 &57.1 &37.1\\
   4&  VBD+spatial&36.1&56.8&37.0\\
   5&  VBD+spatial+semantic&37.9&58.1&38.3\\
   6&  VBD+visual+semantic&39.8 &59.3& 40.0\\
   7&  VBD+visual+spatial&38.5& 58.0& 38.5\\
   8&  VBD+spatial+semantic+visual&\textbf{40.5}&\textbf{60.6}&\textbf{41.9}\\
    \bottomrule
  \end{tabular}
  }
  
  \label{tab:KA}
 \end{table}

\textbf{Different  knowledge aggregation variants.}
Our exploration into the influence of various knowledge aggregation approaches on active object detection has shed light on nuanced improvements, as shown in Table\ref{tab:KA}. Beginning with the baseline, represented in the first row, which exclusively employs the Vision-Based Detector. The introduction of the Knowledge-Enhanced Detector combined with an oracle query containing solely the ground-truth normalized box of the active object showcased in the second row  (VBD+spatial), demonstrates an initial boost in performance (on AP +0.2\% improvement). This indicates that leveraging positional information of active object is indeed beneficial and provides valuable cues for detection.
We delve deeper into enriching the knowledge aggregation process. In the third row(VBD+spatial+semantic), we take a significant step forward by incorporating semantic features of possible interactions. The outcome is a further improvement in detection performance(on AP +1.8\% improvement). This noteworthy progress underscores the pivotal role of semantic information tied to active objects. 
Furthermore, we provide more direct image information for these interactions as part of the oracle query (results in the last row, VBD+spatial+semantic+visual). The visual features provide the best performance(on AP +2.6\% improvement). 
The comparisons show the necessity of triple knowledge: visual-assited,semantic-aware and spatial-sensitive.

\begin{table}
  \centering
  \setlength{\abovecaptionskip}{0cm} 
  \setlength{\abovecaptionskip}{0cm} 
        \setlength{\belowcaptionskip}{-0.2cm}
  \caption{Ablation of knowledge distillation on Ego4D.  }
        \scalebox{0.9}
{
  \begin{tabular}{@{}c|c|ccc@{}}
    \toprule
    {No.}&{Distillation}& AP&AP50&AP75\\
    \midrule
   1&  VBD &{35.9}&{55.8}&{36.9}\\
   2&  VBD $\textit{w}\ {emb}$&{38.3} &59.3&41.2\\
   3&  VBD $\textit{w}\ emb\&attn $&\textbf{40.5}&\textbf{60.6}&\textbf{41.9}\\
    \bottomrule
  \end{tabular}
  }
   
  \label{tab:TSD}
 \end{table}

\textbf{Ablation of different knowledge distillation strategy.}
Additionally, our investigation extended to the realm of different distillation techniques and their influence on detection outcomes in Table~\ref{tab:TSD}. Similar to our earlier explorations, the baseline model(VBD, Vision-Based Detector) is initiated without the incorporation of knowledge aggregation and distillation. Subsequently, we introduced the Knowledge-Enhanced Detector as an extension of the baseline,  with distillation exclusively applied to features. The outcomes reveal that feature-level distillation yields a discernible performance boost with an average enhancement of 2.4\% on AP. This underscores the potential of leveraging feature distillation to foster the acquisition of detection capabilities by the student model (Vision-Based Detector) from the teacher model (Knowledge-Enhanced Detector). Building upon this foundation, the introduction of distillation across attentions imparts a substantial augmentation to the model's proficiency, underscoring the synergistic benefits of comprehensive distillation techniques with an average enhancement of 2.2\% on AP. We  introduce distillation across attention mechanisms in addition to feature distillation. The outcomes of this approach yielded significant advancements in the model's proficiency. This underscores the synergistic potential of comprehensive distillation strategies that not only align features but also bridge the gap between attentions. By orchestrating the transfer of intermediate outputs and attention maps from the teacher to the student, our distillation scheme enables the Vision-Based Detector to harness the enhanced knowledge of the Knowledge-Enhanced Detector.

\textbf{Ablation of the number of generated descriptions.}
We validate the performance of the model using different numbers of semantic descriptions. We generated 10 interaction descriptions of  a state-change object using GPT, with the prompt ``describe 10 interaction descriptions of \textit{[object]} undergoing state change (including tools)".
From Table \ref{tab:text}, it can be seen that diverse descriptions bring significant improvements to model performance(when using 10 text descriptions, detection performance improved by 3.2\% on AP). This may be due to the fact that the scene of a state change of an object may be diverse, so diverse descriptions are necessary.

\textbf{Ablation of the number of generated images .}
We also validate the impact of using different numbers of generated images on model performance, as shown in Table ~\ref{tab:images}.  It can be seen that as the number of generated images increases, the performance of the model also increases. Compared to not using images, when each object uses 1 generated image (randomly selecting a text description to generate one image), the model improves by 0.2\% on AP. 
When the number increased to 10 (10 descriptions for each object and each text description generated 1 image), the model improved by 1.6\% on AP. 
At 100 images (each text description generated 10 images), the model performance provided 2.6\% on AP. This indicates that diverse visual features can provide more performance improvements to the model.

\textbf{Different aggregation approaches.}
In Table \ref{tab:agg}, we validate the impact of aggregation methods({attentive indicates use the way described in Eq.\ref{eq:ag}}) on model performance. In the first two rows, we directly perform max- or average-pooling on semantic features or visual features without any attention operation. It can be seen that the performance of maximum pooling is relatively high(+0.1\% on AP). Furthermore, we first perform a self attention operation on the features and then max-pooling (the third line, attentive). Attentive operation has brought about 1.3\% improvement on AP, which shows adaptive selection contributes to AOD.

\begin{table}
  \centering
  \setlength{\abovecaptionskip}{0cm} 
  \setlength{\abovecaptionskip}{0cm} 
        \setlength{\belowcaptionskip}{-0.2cm}
  \caption{Different  number of generated interaction descriptions.  }
        \scalebox{0.9}
{
  \begin{tabular}{@{}c|c|ccc@{}}
    \toprule
    {No.}&{Number of descriptions}& AP&AP50&AP75\\
    \midrule

   1&  No-description&37.3&57.8&37.7\\
   2&   1-description &37.5&57.9&37.7\\
   3&10-descriptions&\textbf{40.5}&\textbf{60.6}&\textbf{41.9}\\
    \bottomrule
  \end{tabular}
  }
   
  \label{tab:text}
 \end{table}

\begin{table}
  \centering
  \setlength{\abovecaptionskip}{0cm} 
  \setlength{\abovecaptionskip}{0cm} 
        \setlength{\belowcaptionskip}{-0.2cm}
  \caption{Different  number of generated images.  }
        \scalebox{0.9}
{
  \begin{tabular}{@{}c|c|ccc@{}}
    \toprule
    {No.}&{Number of generated images}& AP&AP50&AP75\\
    \midrule

   1&  No-image&37.9&58.1&38.3\\
   2&   1-image &38.1&58.2&38.4\\
   3&10-images&39.5&58.7&39.1\\
   4&100-images&\textbf{40.5}&\textbf{60.6}&\textbf{41.9}\\
    \bottomrule
  \end{tabular}
  }
   
  \label{tab:images}
 \end{table}

 \begin{table}
  \centering
  \setlength{\abovecaptionskip}{0cm} 
   \setlength{\abovecaptionskip}{0cm} 
        \setlength{\belowcaptionskip}{-0.2cm}
  \setlength{\abovecaptionskip}{0cm} 
        \setlength{\belowcaptionskip}{-0.2cm}
  \caption{Different  aggregation approaches.  }
        \scalebox{0.9}
{
  \begin{tabular}{@{}c|c|ccc@{}}
    \toprule
    {No.}&{method}& AP&AP50&AP75\\
    \midrule

   1&  max &39.2&59.5&39.6\\
   2&    avg &39.1&59.2&39.7\\
   3&attentive&\textbf{40.5}&\textbf{60.6}&\textbf{41.9}\\
    \bottomrule
  \end{tabular}
  }
   
  \label{tab:agg}
 \end{table}

\section{Conclusion}
\label{sec:conclusion}
We aim to address the challenges inherent in active object detection by knowledge aggregation and distillation.  We propose a framework that significantly improves the accuracy and efficiency of active object detection.
Our proposed Knowledge Aggregator aggregates three-fold commonsense pertaining to active objects, encompassing plausible semantic interactions, fine-grained visual and spatial priors. Furthermore, our Knowledge Distillation strategy empowers the traditional detector with the capability for localizing active objects without extra prior inputs.
The results of comprehensive experiments on  Ego4D, Epic-Kitchens, 100DOH and MECCANO,  demonstrate the efficacy of our method. 

\noindent \textbf{Acknowledgements.}
This work was supported by the grants from the National Natural Science Foundation of China 62372014.

{\small
\bibliographystyle{ieeenat_fullname}
\bibliography{11_references}
}

\end{document}